\documentclass[letterpaper, 10 pt, conference]{ieeeconf}  

\IEEEoverridecommandlockouts                              

\usepackage{graphics}
\usepackage{epsfig} 
\usepackage{mathptmx} 
\usepackage{times} 
\usepackage{amsmath} 
\usepackage{amssymb}  
\usepackage{multirow}
\usepackage{url}
\usepackage{caption}
\usepackage{subcaption}
\usepackage[table]{xcolor}
\usepackage{import}
\usepackage{hyperref}
\usepackage{booktabs}
\usepackage{tabularx}
\usepackage{microtype}
\usepackage{gensymb} 
\usepackage{cite}
\usepackage{stfloats}

\title{\LARGE \bf
Static-Equilibrium Oriented Interaction Force Modeling and Control of Aerial Manipulation with Uni-Directional Thrust Multirotors
}

\author{ Tong Hui, Matteo Fumagalli
\thanks{This work has been supported by the European Unions Horizon 2020 Research and Innovation Programme AERO-TRAIN under Grant Agreement No. 953454.}
\thanks{The authors are with Department of Electrical and Photonics Engineering, Technical University of Denmark
        {\tt\small tonhu@dtu.dk}
        {\tt\small mafum@dtu.dk}}
}

\begin{document}

\maketitle
\thispagestyle{empty}
\pagestyle{empty}

\begin{abstract}

This paper presents a static-equilibrium oriented interaction force modeling and control approach of aerial manipulation employing uni-directional thrust (UDT) multirotors interacting with variously defined environments. First, a simplified system model for a quadrotor-based aerial manipulator is introduced considering parameterized work surfaces under assumptions, and then a range of meaningful manipulation tasks are utilized to explore the system properties in a quasi-static equilibrium state. An explicit interaction force model in relation with the aerial manipulator pose configuration and the environment parameter is derived from the static equilibrium analysis, based on which singularity is pointed out. Then a hybrid attitude/force interaction control strategy is presented to verify the proposed interaction force model, which involves high gain attitude control and feedforward plus feedback force control. This paper represents  preliminary results. We study the properties of UDT-based aerial manipulators via specific tasks, and propose a novel framework for interaction force modeling and control aiming at maximizing the commercial values of UDT platforms for aerial manipulation purpose.

\end{abstract}

\section{INTRODUCTION}
The field of aerial manipulation has shown sustained growth during the last decade towards developing great physical interaction capabilities for industrial needs \cite{Anibal2021}. Multirotor platforms are widely used for aerial manipulation purpose and among which, the Uni-Directional Thrust (UDT) platforms are the most common commercially available products, e.g. quadrotor, hexarotor \cite{Anibal2021},\cite{Hamandi2021}. However, these UDT-based aerial manipulation systems often suffer from limited interaction capabilities due to underactuation of the aerial vehicle \cite{Park2018}. This paper thus explores the wider range of employing UDT-based aerial manipulators to physically interact with variant environments and complete required manipulation tasks by proposing a static-equilibrium based approach for interaction force modeling and control.

In the literature, some manipulation tasks are already addressed experimentally, such as contact-based inspection \cite{Tognon2019,Bodie2021}, grasping \cite{Mellinger2011}, torsional manipulation \cite{Shimahara2016,Korpela2014}, operating with movable structure \cite{Kim2015,Door2015,Lee2021}, drilling/screwing \cite{Ding2021} and etc., and among which only a few are employed for real industrial usage. These tasks require physical interaction between the environment and the system which involves an exchange of forces/torques \cite{Anibal2021}. Moreover, variant physical interaction cases require the system to exert desired forces/torques accordingly in sense of both direction and magnitude which may exceed the ability of the aerial manipulator.

One way to bring aerial manipulation widely into industrial applications is to enhance the capability of generating interaction forces in multiple directions \cite{Park2018}, \cite{Ryll2019}. For but not limited to this purpose, UDT multirotors with a high degree of freedom (DoF) manipulator attached \cite{Fuma2014,Suarez2020}, \cite{Suarezf2018} and aerial vehicles with multi-directional thrust (e.g. omnidirectional platforms \cite{Park2018}, \cite{Bodie2019}, \cite{Voliro} including fully actuated aerial vehicle \cite{Franchi2018}) came into the picture of aerial manipulation. The high degree of freedom manipulators often add control complexity and energy consumption to the system even with light weight design as in \cite{Suarezf2018}. The interests of employing multi-directional thrust platforms for aerial manipulation purpose are rising in the community recently and these platforms often allow the decoupling between translation and rotation dynamics of the system. However, they are often much more expensive due to mechanical and actuation complexity compared with the UDT multirotor products and not widely available in the market yet to the best of the authors' knowledge. The major drone market still focuses on UDT multirotors which makes it valuable to further investigate in the feasibility of using these platforms for variant manipulation tasks as in \cite{PL2022} and also motivates the work of this paper. This paper thus studies a UDT-based aerial manipulator with an one DoF manipulator attached.

The interaction control of aerial manipulation determines the functionalities of the aerial manipulator towards potential industrial applications \cite{Anibal2021}. One of most common control techniques to handle interaction forces is the impedance/admittance control \cite{Anibal2021}, \cite{Nava2020}. In \cite{Ryll2019}, this strategy is applied on a fully actuated aerial vehicle, and in \cite{Fumagalli2012}, it is employed on an underactuated platform. This technique, however, makes the system behave rather passive while interacting with the environment \cite{Lee2013} and instead, another common control technique - hybrid position/force control strategy allows the system to actively exert desired interaction force, as in 
\cite{Bodie2021}, \cite{Nava2020}, \cite{Lee2013}, \cite{meng2019}. To have accurate force control, obtaining force information via estimation \cite{Tomic2017} or direct measurements from force sensors \cite{Nava2020} are required, moreover, contact model has to be properly addressed which results in system constraints. Actively controlling the interaction forces/torques during physical interaction is closely relevant to expanding the range of executable manipulation tasks with aerial robots, however, it is often difficult to control both force magnitude and direction using underactuated aerial vehicles like UDT multirotors while having stable contact with the environment and motion constraints. The studies conducted in \cite{HW2017}, \cite{PL2022} represent early efforts in investigating the interplay between UAV attitude, gravity, thrust, and exerted forces in UDT multirotor-based aerial manipulators to generate continuous forces while controlling the UAV attitude and contact position. It is evident that for UDT multirotors, the capability of generating forces/torques is closely related to the system weight, attitude of the UAV and thrust saturation due to the coupled translation and rotation dynamics. This paper therefore enhances the existing research by investigating a range of manipulation tasks and proposing viable modeling and control approaches for physical interaction, which effectively leverage the inherent coupling properties of UDT-based aerial manipulators.

Considering now only the interaction tasks with a static environment and a single contact point \cite{Nava2017,Ale1994}, the work surface with which the aerial manipulator interacts is assumed to be rigid and flat, and is often considered as vertical \cite{Ding2021} or horizontal \cite{kim2020}. In reality, the orientation of the work surface can be variously determined by the industrial application cases. This paper investigates the feasibility of utilizing UDT platforms for aerial manipulation tasks, specifically in the context of interacting with various work surfaces, moreover, a static-equilibrium analysis based interaction force modeling and control framework is proposed in this paper. In \cite{park2020}, an equilibrium-based strategy is introduced. However, our work presents a straightforward interaction force model that is explicitly expressed with respect to the robot pose configuration and environmental parameters. This model is derived through a static equilibrium analysis of the entire system during stationary interaction with the environment. A hybrid attitude/force control technique is then proposed based on the aforementioned interaction force model.

 The paper is organized as follows. Sec.~\ref{sec:bg} presents the system modeling and problem statement. Sec.~\ref{sec:pi} introduces the static equilibrium orientated interaction force modeling approach and singularity analysis. Sec.~\ref{control} presents the control strategy to verify the proposed interaction force model. Sec.~\ref{sec:simu} displays the simulation results of the applied control strategy for physical interaction. Sec.~\ref{sec:con} concludes the paper and points out limits and future work.

\section{Background}
\label{sec:bg}

\subsection{Simplified System Modeling}
\label{sec:notation}
To study the capability of actively exerting desired interaction wrenches for aerial manipulation purpose with UDT-based aerial manipulators, we propose a range of contact-based manipulation tasks which allows us to explore the underactuated system with coupled translation and rotation dynamics. Consider an Unmanned Aerial Manipulator (UAM) composed of a quadrotor equipped with a 1-DoF manipulator/link connected by an actuated joint to interact with a flat work surface. 

\begin{figure}[t]
      \centering
      \includegraphics[width=\columnwidth]{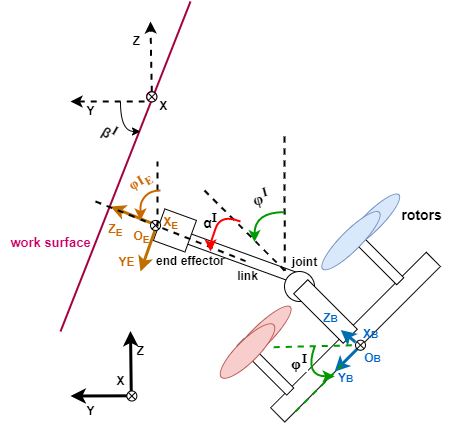}
      \caption{Simplified System Modeling}
      \label{fig:ad}
\end{figure}

Define the inertial frame $\mathbb{I}=\{O;X,Y,Z\}$ as a reference frame of any motion. Let $\boldsymbol{B}=\{O_B;X_B,Y_B,Z_B\}$ be the body frame attached to the center of gravity of the aerial vehicle. We assume the geometric center of the aerial vehicle is its center of gravity (CoG). Let $\boldsymbol{p_B^\mathbb{I}} \in \boldsymbol{R}^3$ represent the relative position of $O_B$ of the body frame $\boldsymbol{B}$ w.r.t. the inertial frame $\mathbb{I}$ expressed in frame $\mathbb{I}$. Let $\boldsymbol{E}=\{O_E;X_E,Y_E,Z_E\}$ be the end-effector (e.e.) frame attached to the end-effector tip of the manipulator which is also the contact location during physical interaction. To simplify the problem while maintaining fundamental contribution of this work, the system modeling is restricted to a two dimensional (2-D) representation as show in Fig.~\ref{fig:ad} considering the following assumptions:  
\begin{itemize}
    \item \textbf{Assumption 1:} the rotation axis of the actuated joint is parallel to the axis $\boldsymbol{X}_B$ of the body frame $\boldsymbol{B}$, moreover, the actuated joint locates on the axis $\boldsymbol{Z}_B$, i.e. the manipulator's placement is limited inside the plane $(\boldsymbol{Y}_B, \boldsymbol{Z}_B)$; 
    \item \textbf{Assumption 2:} the orientation of the body frame $\boldsymbol{B}$ is restricted inside the plane $(\boldsymbol{Y}, \boldsymbol{Z})$ of the inertial frame around the axis $\boldsymbol{X}$.
\end{itemize} 
The work surface is obtained by rotating the plane $(\boldsymbol{X},\boldsymbol{Y})$ of the inertial frame around $\boldsymbol{X}$ axis with an angle $\beta^{\mathbb{I}}$, where $\beta^{\mathbb{I}} \in [-\frac{\pi}{2},\frac{\pi}{2}]$ being positive when it rotates anticlockwise around positive $\boldsymbol{X}$ of the inertial frame. The work surface is parameterized by $\beta^{\mathbb{I}}$ to simulate possible industrial environment conditions. 
  Let the e.e. frame $\boldsymbol{E}$ have the same orientation as the body frame $B$ when $\alpha^{\mathbb{I}}=0$, where we define the orientation angle from axis $\boldsymbol{Z}_B$ to axis $\boldsymbol{Z}_E$ expressed in the inertial frame $\mathbb{I}$ as the joint position $\alpha^{\mathbb{I}}$. Define the orientation of the body frame around the axis $\boldsymbol{X}$ w.r.t. the inertial frame as the roll angle $\varphi^{\mathbb{I}}$ expressed in the inertial frame. And $\alpha^{\mathbb{I}}, \varphi^{\mathbb{I}} \in [-\frac{\pi}{2},\frac{\pi}{2}]$ are positive while rotating anticlockwise around positive $\boldsymbol{X}$ axis. The orientation of the e.e. frame $\boldsymbol{E}$ w.r.t the inertial frame in the restricted plane can then be presented as: 
\begin{equation}
\varphi_E^{\mathbb{I}}=\varphi^{\mathbb{I}}+\alpha^{\mathbb{I}},
\label{eq:eeori}
\end{equation}
expressed in the inertial frame. Let $\boldsymbol{R}_B(\varphi^{\mathbb{I}}), \boldsymbol{R}_E(\varphi_E^{\mathbb{I}}) \in SO(3)$ denote the associated rotation matrices of the body frame $\boldsymbol{B}$ and the e.e. frame $\boldsymbol{E}$ w.r.t. the inertial frame respectively.

Consider $\boldsymbol{F}_C^{\mathbb{I}}=\begin{bmatrix}
 \boldsymbol{f}_C^{\mathbb{I}}\\ \boldsymbol{\tau}_C^{\mathbb{I}}
\end{bmatrix} \in \boldsymbol{R}^6$ as the interaction wrenches exerted from the end-effector of the UAM acting on the environment expressed in the inertial frame, assuming that the interaction wrenches at the end-effector tip are the only external forces/torques exchange between the environment and the system.

\begin{figure}[t]
      \centering
     \includegraphics[width=0.5\columnwidth]{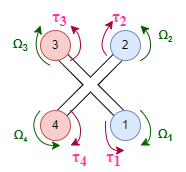}
      \caption{propeller motor layout}
      \label{fig:motor}
\end{figure}

The layout and rotating direction of the propellers are shown in Fig.~\ref{fig:motor}. Define $\Omega_i$ as the propeller speed of the $i$ th propeller, where $i=1,2,3,4$. Let $T_i$ and $\tau_i$ present the magnitude of the thrust and aerodynamic moment respectively generated by the $i$ th propeller. 
 $\mathbf{T}_i^{\mathbb{I}} \in \boldsymbol{R}^3$ presents the thrust vector of the $i$ th propeller expressed in the inertial frame $\mathbb{I}$, and  $\Gamma_i^{\mathbb{I}} \in \boldsymbol{R}^3$ presents the torque vector generated by the $i$ th propeller w.r.t. the CoG of the quadrotor expressed in the inertial frame $\mathbb{I}$, then one has:

\begin{equation}
\begin{split}
    T_i&=k_{af}\Omega_i^2,\\
    \tau_i&=k_{am}\Omega_i^2,\\
    \mathbf{T}_i^{\mathbb{I}}&=\boldsymbol{R}_B (T_i\cdot \begin{bmatrix}
     0\\0\\1
    \end{bmatrix}), \\
    \Gamma_i^{\mathbb{I}}&=\boldsymbol{R}_B\Big(\boldsymbol{r}_i^B\times (T_i\cdot \begin{bmatrix}
     0\\0\\1
    \end{bmatrix})+(-1)^{i+1}\tau_i\cdot \begin{bmatrix}
     0\\0\\1
    \end{bmatrix}\Big),\\
    i&=1,2,3,4.
\end{split}
\end{equation}
where $k_{af},k_{am}$ are respectively the aerodynamic coefficients of thrust and moment w.r.t. $\Omega_i^2$. $\boldsymbol{r}_i^B$ is the coordinate vector of the center of the $i$ th propeller w.r.t. the origin of the body frame $\boldsymbol{B}$ expressed in frame $B$, for more details, refer to \cite{Ding2021}.

\subsection{Problem Statement}
\label{sec:ps}
In this section, interacting with variously orientated work surfaces w.r.t. the inertial frame is considered in the task requirements. To explore the feasibility of applying designated interaction forces/torques on the environment using UDT-based aerial manipulators while having stationary contact with the parameterized work surfaces, the UAM system introduced in the previous section is required to excute the following tasks:
\begin{itemize}
    \item \textbf{Task} 1) the end-effector is orientated such that the axis $\boldsymbol{Z}_E$ of the e.e. frame is orthogonal to the work surface;
    \item \textbf{Task} 2) maintain stationary contact while interacting with the work surface;
    \item \textbf{Task} 3) exert a linear interaction force vector $\boldsymbol{f}_C^{\mathbb{I}}$ to the environment along with the positive axis $\boldsymbol{Z}_E$ of the e.e. frame continuously, which is known as a pushing task with a single contact point at the tip of the end-effector.
\end{itemize}
The aforementioned three tasks serve as preliminary steps towards facilitating various significant industrial manipulation tasks, including but not limited to drilling and screwing. \cite{Ding2021}, Peg-in-Hole assembly \cite{car2018}. The detailed characterization of rotational interaction torques for complex operations, such as drilling, extends beyond the scope of this paper and warrants further investigation.

Considering the aforementioned 2-D presented system, \textbf{Task} 1) introduces constraints on the e.e.'s orientation which can be represented by $\beta^{\mathbb{I}}=\varphi_E^{\mathbb{I}}$ assuming that the aerial manipulator only operates below the work surface (i.e. $\varphi_E^{\mathbb{I}}$ has the same sign as $\beta^{\mathbb{I}}$). With Eq.~(\ref{eq:eeori}) the below relation is introduced: 
\begin{equation}
\beta^{\mathbb{I}}=\varphi^{\mathbb{I}}+\alpha^{\mathbb{I}}.
 \label{eq:angle}
\end{equation}
The relation holds for both positive and negative $\beta^{\mathbb{I}}$ defined work surface.  \textbf{Task} 2) implements the zero motion constraints to the end-effector tip at the contact point as stated in \cite{Nava2017} which results in zero motion of the aerial vehicle by kinematic transformation, in the meanwhile \textbf{Task} 3) requires desired interaction forces expressed in the e.e. frame $\boldsymbol{E}$ as $\boldsymbol{F}_C^E$ to be in the form of:
 
\begin{equation}
\label{eq:force_req}
     \begin{split}
         &\boldsymbol{F}_C^E=\begin{bmatrix}
        0&0&\boldsymbol{f}_{C,Z_E}^*&\boldsymbol{0}_3
         \end{bmatrix}^{\top},\\
         &\boldsymbol{f}_{C,Z_E}^*=\begin{bmatrix}
    0&0&1
\end{bmatrix}\cdot(\boldsymbol{R}_E^{\top}\boldsymbol{f}_C^{\mathbb{I}}) \geq0.
     \end{split}
 \end{equation}
where $\boldsymbol{R}_E^{\top}\boldsymbol{f}_C^{\mathbb{I}}$ is the desired linear interaction force vector expressed in the e.e. frame $\boldsymbol{E}$ at the e.e. tip.

 With the introduced UAM system and work surfaces for the required manipulation tasks, it is now possible to analyze the quasi-static equilibrium condition of the system during stationary physical interaction.
 \section{Physical Interaction}
\label{sec:pi}

\subsection{Static-Equilibrium Analysis}
\label{sec:force model}

\begin{figure}[t]
      \centering
      \includegraphics[width=\columnwidth]{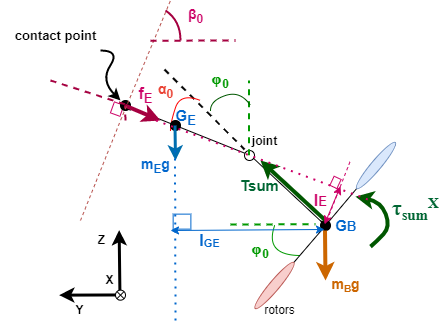}
      \caption{Free Body Diagram}
      \label{fig:fbd}
\end{figure}

Neglecting the dynamics of the propeller motors and the joint servo motor and assuming that the actuator at the joint is sufficient to always hold the attached manipulator in place, during the stationary physical interaction described in Sec.~\ref{sec:ps}, the whole system can be considered as a rigid body resting at a static equilibrium phase with zero net forces and torques to maintain the equilibrium both in translation and rotation while interacting with the surface. Thus a simplified free body diagram (FBD) of the UAM system in the plane $(\boldsymbol{Y},\boldsymbol{Z})$ of the inertial frame during the interaction under a certain robot pose configuration $(\varphi^{\mathbb{I}},\alpha^{\mathbb{I}})$ and a defined work surface can then be displayed as in Fig.~\ref{fig:fbd}, in which we consider $\beta_0=|\beta^{\mathbb{I}}|,\varphi_0=|\varphi^{\mathbb{I}}|,\alpha_0=|\alpha^{\mathbb{I}}|$.

 $\boldsymbol{G}_E$ is the CoG of the manipulator and $\boldsymbol{G}_B$ is the CoG of the aerial vehicle. $m_B,m_E$ are the mass of the aerial vehicle and the manipulator respectively. $l_E$ is the minimum distance between the desired contact force vector acting on the contact point and the CoG of the aerial vehicle $\boldsymbol{G}_B$, and $l_{G_E}$ is the minimum distance between the gravity force vector of the manipulator acting on $\boldsymbol{G}_E$ and $\boldsymbol{G}_B$. Considering \textbf{Assumptions 1\&2}, one has $T_{sum}=\sum_{i=1}^4 T_i$ which is the total thrust magnitude generated by 4 propellers and $\tau_{sum}^X=|\begin{bmatrix}
     1&0&0
    \end{bmatrix}\cdot(\sum_{i=1}^4 \boldsymbol{\Gamma}_i^{\mathbb{I}})|$ which is the total rotational torque magnitude around axis $\boldsymbol{X}$ of the inertial frame caused by 4 propellers, while $\tau_{sum}^Y=\tau_{sum}^Z=0$. We apply $f_E=|-f^*_{C,Z_E}|$ as the desired contact force magnitude acting on the system from the environment along the axis $\boldsymbol{Z}_E$ of the e.e. frame $\boldsymbol{E}$. Considering $\boldsymbol{G}_B$ as the reference point, the forces/torques acting on the system at the equilibrium state are thus displayed below.

\begin{itemize}
    \item Linear Forces along $\boldsymbol{Z}$ axis of the inertial frame:
\begin{equation}
    T_{sum}\cdot cos(\varphi_0)=m_Eg+m_Bg+f_E \cdot cos(\beta_0),
    \label{eq:z}
\end{equation}

   \item Linear Forces along $\boldsymbol{Y}$ axis of the inertial frame:
\begin{equation}
    T_{sum} \cdot sin(\varphi_0)=f_E \cdot sin(\beta_0),
    \label{eq:y}
\end{equation}

  \item Rotational Torques around $\boldsymbol{X}$ axis of the inertial frame:
\begin{equation}
    \tau_{sum}^X +m_Eg \cdot l_{G_E}=f_E \cdot l_E.
    \label{eq:torque}
\end{equation}
\end{itemize}
With the focus of linear forces acting on the system and the following assumptions:
\begin{itemize}
    \item $\alpha_0 \neq 0$, $\beta_0 \neq 0$,
    \item $\varphi_0 < \beta_0$,
    \item $\beta^{\mathbb{I}}, \varphi^{\mathbb{I}}, \alpha^{\mathbb{I}}$ always have the same sign which ensures that the relation $\beta_0=\varphi_0+\alpha_0$ holds,
\end{itemize}
by re-arranging the Eq. (\ref{eq:z}) (\ref{eq:y}), 
 one has:
\begin{align}
    &f_E=G_t\frac{sin(\varphi_0)}{sin(\alpha_0)},
    \label{eq:zforce}\\
    &f_E^Z=f_E \cdot cos(\beta_0)=G_t\frac{sin(\varphi_0)cos(\beta_0)}{sin(\alpha_0)},
    \label{eq:f_E_Z}\\
    &T_{sum}=G_t\frac{sin(\beta_0))}{sin(\alpha_0)},
    \label{eq:tsum}
\end{align}
where $G_t=m_Eg+m_Bg$, and $f_E^Z$ is the desired interaction force magnitude along negative axis $\boldsymbol{Z}$ of the inertial frame 

Thus, for any fixed environment parameter $\beta_0$ with $\alpha_0\neq 0$ (i.e. $\beta_0-\varphi_0\neq 0$), the interaction force magnitude and the corresponded total thrust magnitude required by the tasks described in Section \ref{sec:ps} is a function of the defined roll angle magnitude $\varphi_0$ of the aerial vehicle and environment parameter $\beta_0$ multiplied by the total gravity force of the system. 

While theoretically, a state of dynamic equilibrium can also be achieved by applying zero net forces and torques, resulting in the object moving at a constant velocity, however, this state is not the focus of this paper. 

\subsection{Singularity Analysis}
\label{sec:singu}

\begin{figure}[t]
      \centering
      \includegraphics[width=\columnwidth,height=5cm]{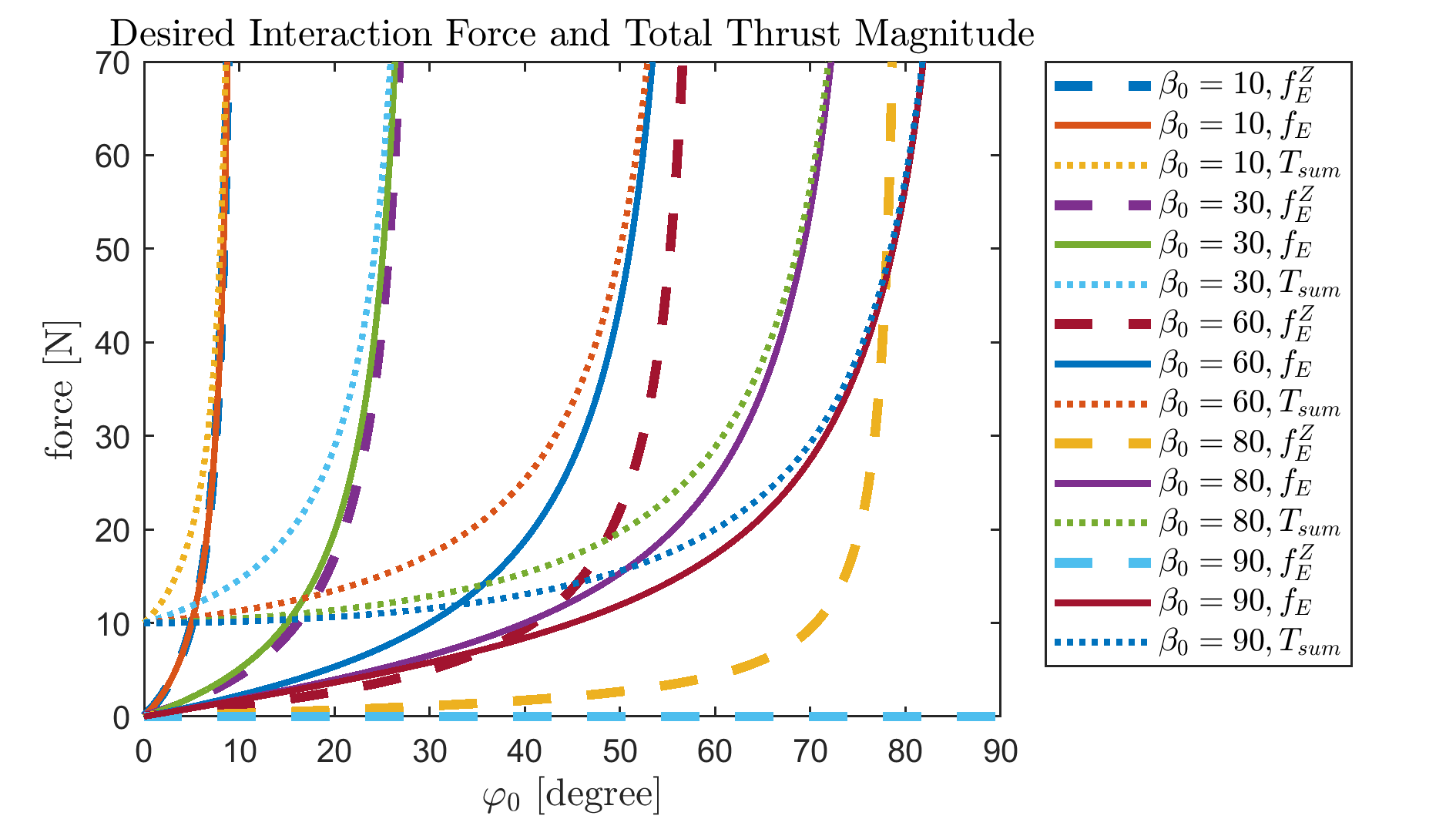}
      \caption{force profiles of Eq.~(\ref{eq:zforce})(\ref{eq:f_E_Z})(\ref{eq:tsum}) for: $\beta_0=10\degree, 30\degree, 60\degree,80\degree, 90\degree$ with $\varphi_0 \in [0, \beta_0)$.}
      \label{fig:plotf}
\end{figure}

The singularity analysis is based on the mathematical expressions of the force models in Eq. (\ref{eq:f_E_Z}), (\ref{eq:f_E_Z}), (\ref{eq:tsum}) derived from the static equilibrium condition. Considering the symmetric definition of the angles and the platform geometry, force models are analyzed for $\beta_0=|\beta^{\mathbb{I}}| \in (0\degree,90\degree]$. By visualizing the properties of the proposed force models in Fig.~\ref{fig:plotf} for $\beta_0=10\degree, 30\degree, 60\degree, 80\degree, 90\degree$, the force profiles show that: by knowing a fixed value of environment parameter $\beta_0$, there is a unique value of the desired interaction force magnitude and total thrust magnitude corresponding to a certain roll angle magnitude $\varphi_0$. The contact force magnitude increases along with the roll angle magnitude and so as the total thrust magnitude required to generate the desired interaction force. Moreover, a small variation of the roll angle magnitude causes large variation of the interaction force and total thrust magnitude when $\varphi_0$ gets close to $\beta_0$. The contact force and total thrust magnitude goes to infinity when $\varphi_0=\beta_0$, i.e. $\alpha_0=0$, which is a singularity point. If and only if $\beta_0=0$, the platform is capable of generating force along $\boldsymbol{Z}_E$ of the e.e. frame being perpendicular to the work surface together with $\alpha_0=0$ and $\varphi_0=0$, and this force magnitude is limited by the propeller thrust saturation, known as operating towards the ceiling \cite{lan2022} or operating downwards \cite{kim2020}.

In the next section, a control strategy for the studied problem is designed to verify the proposed static-equilibrium orientated interaction force model.

\section{Control Design}
\label{control}
\begin{figure*}[t]
      \centering
      \includegraphics[width=0.8\textwidth]{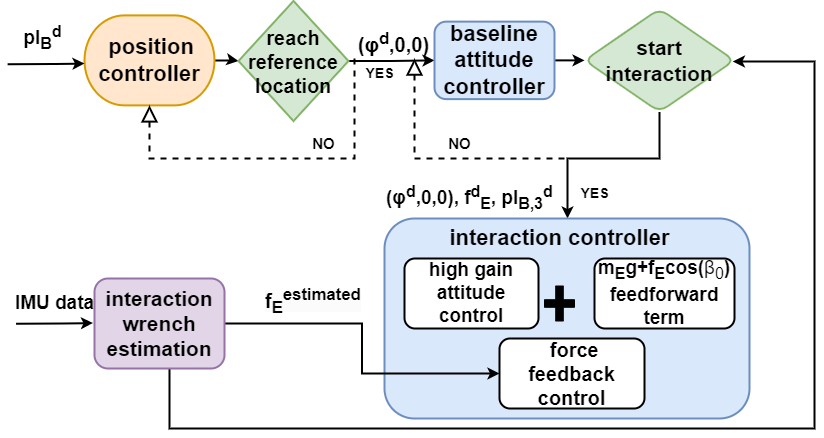}
      \caption{Control Strategy: Work Flow }
      \label{fig:wf}
\end{figure*}

The baseline controller of the aerial vehicle contains a cascaded position controller and a geometric attitude controller (for details, refer to \cite{Lee2011}), and the baseline attitude controller generates system inputs as desired moments $\begin{bmatrix}
    M_X&M_Y&M_Z
\end{bmatrix}^{\top}$ based on the attitude dynamics of the aerial vehicle and the desired total thrust magnitude $T_{sum}^{base}$ to maintain the altitude (position along $\boldsymbol{Z}$ axis of the inertial frame) of the aerial vehicle as below:

\begin{equation}
    \begin{split}
    T_{sum}^{base}=&\big[k_p(p_{B,3}^{\mathbb{I}}-p_{B,3}^{\mathbb{I}d})+k_d(\dot{p}_{B,3}^{\mathbb{I}}-\dot{p}_{B,3}^{\mathbb{I}d})\\
    &-m_B\ddot{p}_{B,3}^{\mathbb{I}d}+m_Bg\big]\frac{1}{\boldsymbol{e}_3 \cdot \boldsymbol{R}_B \boldsymbol{e}_3}
    \end{split}
    \label{eq:tsum_base}
\end{equation}
where $k_p$ and $k_{\upsilon}$ are the positive definite gains and $p_{B,3}^{\mathbb{I}d}$ is the desired altitude of the aerial vehicle. We assume that $\boldsymbol{e}_3 \cdot \boldsymbol{R}_B \boldsymbol{e}_3 \neq 0$, see \cite{Lee2011}.

A control strategy as in Fig.~\ref{fig:wf} is designed for physical interaction tasks presented in Sec.~\ref{sec:ps}. With the static-equilibrium based interaction force modeling, the interaction force magnitude along desired direction is thus directly related to the aerial vehicle attitude (i.e. the roll angle) which makes it feasible to achieve the objectives via a hybrid attitude/force control. The designed interaction controller combines a high gain attitude control together with a feedforward plus feedback force control.
 
\subsection{High Gain Attitude Control}
To generate a reference interaction force with a magnitude of 
\begin{equation}f_E^d=G_t\frac{sin(\varphi_0^d)}{sin(\alpha_0^d)}
\end{equation}
along the desired direction on the defined work surface, the joint position magnitude of the manipulator is then determined by $\alpha_0^d=\beta_0-\varphi_0^d$. And $\alpha^{\mathbb{I},d}=k_s\alpha_0^d,\varphi^{\mathbb{I},d}=k_s\varphi_0^d$, where $k_s=1$ when $\beta^{\mathbb{I}}$ is positive, and $k_s=-1$ when $\beta^{\mathbb{I}}$ is negative. Place the manipulator according to $\alpha^{\mathbb{I},d}$ on the aerial vehicle, the aerial vehicle is then commanded to approach the work surface starting from a reference position\footnote{The reference position to start approaching can be selected according to the reference contact point location, and the motion of the aerial vehicle can be properly planned which is out of the scope of this paper.} with a desired roll angle $\varphi^{\mathbb{I},d}$ (the reference orientation of the UAV along other directions are set as zeros), which enforces the UAM to actively exert forces along the designated direction on the environment according to the proposed interaction force model. High gains on attitude control are thus preferable to avoid evident orientation errors which might lead to large force division w.r.t. the reference interaction force being away from the quasi-static equilibrium state and even bring the operation point close to the singularity point especially while interacting with the work surface defined by small $\beta_0$, based on the analysis in Sec.~\ref{sec:singu}.

\subsection{Feedforward plus Feedback Force Control}
Based on the static-equilibrium analysis in Sec.~\ref{sec:force model}, the disturbances caused by the dynamics of the manipulator $\boldsymbol{D}$ during stationary physical interaction expressed in the inertial frame is: 
\begin{equation}
    \boldsymbol{D}=\begin{bmatrix}
        0\\0\\-m_Eg\\-m_Eg \cdot l_{G_E}\\0\\0
    \end{bmatrix}
    \label{eq:d}
\end{equation}
which involves only the gravity effect of the manipulator at the quasi static-equilibrium state. Similarly, the disturbances caused by interaction forces/torques acting on the CoG of the aerial vehicle expressed in the inertial frame can be displayed as:
\begin{equation}
    \boldsymbol{D}_E=\begin{bmatrix}
        0\\-f_E^d \cdot sin(\beta_0)\\-f_E^d \cdot cos(\beta_0)\\f_E \cdot l_E\\0\\0
    \end{bmatrix}
    \label{eq:de}
\end{equation}

 The third elements of Eq.~(\ref{eq:d}) and (\ref{eq:de}) representing the total external force components along the $\boldsymbol{Z}$ axis of the inertial frame are then added to the altitude control of the baseline attitude controller to enforce the total thrust magnitude for compensating the external forces caused by manipulator gravity and physical interaction with the environment as:
 \begin{equation}
\begin{split}
         &T_{sum}^1=[k_p(p_{B,3}^{\mathbb{I}}-p_{B,3}^{\mathbb{I}d})+k_d(\dot{p}_{B,3}^{\mathbb{I}}-\dot{p}_{B,3}^{\mathbb{I}d})\\
         &-m_B\ddot{p}_{B,3}^{\mathbb{I}d}+m_tg+f_E^d \cdot cos(\beta_0)]\frac{1}{\boldsymbol{e}_3 \cdot \boldsymbol{R}_B \boldsymbol{e}_3},
          \end{split}
\end{equation}
 where $m_t=m_E+m_B$.

For better tracking the exerted interaction forces/torques, a PID feedback control on the force magnitude generated along the desired direction is added to the total thrust as:
\begin{align}
\begin{split}
         &T_{sum}^{int}=T_{sum}^1+\frac{u_f}{\boldsymbol{e}_3 \cdot \boldsymbol{R}_B \boldsymbol{e}_3},\\ 
         &u_f=K_{p,f}(f_E^d-f_E^{estimated})+K_{d,f}(-\dot{f}_E^{estimated})\\
         &+K_{i,f}\int_0^t(f_E^d-f_E^{estimated})d(\tau).
         \end{split}
         \label{eq:tsum_int}
\end{align}
where $K_{p,f}$,$K_{d,f}$,$K_{i,f}$ are positive definite gains, $T_{sum}^{int}$ is the total thrust magnitude to be fed to the UAM system during interaction.
\subsection{Interaction Force Estimation}

 To obtain information of the interaction forces/torques exerted at the e.e. tip, an external wrench observer which only requires IMU data presented in \cite{Tomic2017} is implemented assuming that there are no other external forces acting on the system. The estimated external wrench acting on the CoG of the aerial vehicle thus contains both the effects of the manipulator dynamics and the interaction wrench from the environment. The Eq.(\ref{eq:d}) is then deducted from the estimated external wrench to obtain the interaction forces/torques information assuming that the manipulator has only gravity effects to the aerial vehicle system. The estimated interaction forces/torques expressed in the inertial frame are then subsequently transformed into the end-effector frame. This force information is then used to trigger the switch between the baseline attitude controller and the designed interaction controller by setting a proper threshold (see Fig.~\ref{fig:wf}). In the interaction controller, the estimated interaction force magnitude along the desired direction is then fed to the controller to close the force feedback control loop. The system inputs generated by the interaction controller is thus $\boldsymbol{u}=\begin{bmatrix}
   T_{sum}^{int}&
       \boldsymbol{M}_X&
        \boldsymbol{M}_Y&
         \boldsymbol{M}_Z
   \end{bmatrix}^{\top}$.

\section{Simulation Results}
\label{sec:simu}
\begin{figure}
      \includegraphics[width=\columnwidth]{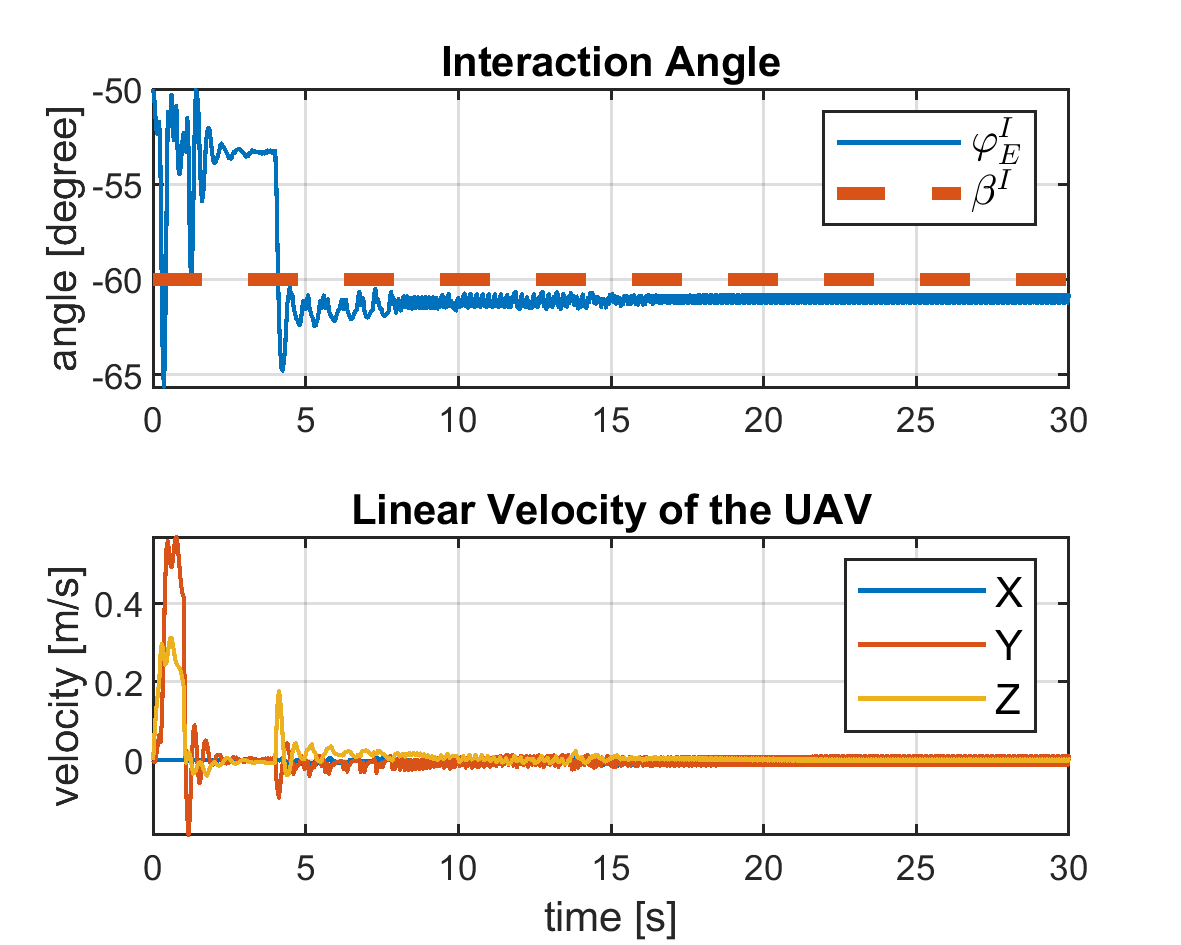}
      \caption{Case 3: $\beta^{\mathbb{I}}=-60 \degree,\varphi^{\mathbb{I},d}=-10\degree$.}
      \label{fig:60_10}
\end{figure}

A Matlab SimMechanis-based quadrotor system equipped with a rigid link with an orientation of $\alpha^{\mathbb{I}}$ as defined in Sec.~\ref{sec:bg} is used to verify the aforementioned interaction force modeling and control framework while the baseline controller is robust enough to maintain stable free flight with the manipulator attached. 
\begin{table}[b]
\begin{center}
\caption{simulation cases}
\begin{tabular}{|c|c|c|c|c|c|} 
 \hline
 Case Num. & $\beta^{\mathbb{I}}  (\degree)$ & $\varphi^{\mathbb{I},d}  (\degree)$ & $\alpha^{\mathbb{I},d}(\degree)$ & $f_E^d$ (N) & $T_{sum}^d$ (N)\\
 \hline
 1  & -30  & -5 & -25 &  1.5370 & 8.8177\\
  \hline
 2  & -30  & -10 & -20 &  3.7840 & 10.8956\\
  \hline
 3 & -60 & -10 & -50 & 1.6895 &  8.4258\\
  \hline
4 & -60 & -15 & -45 &  2.7280 & 9.1281\\
  \hline
 5 & -90 & -15 & -75 &  1.9970 & 7.7160\\
  \hline
 6 & -90 & -20 & -70 & 2.7127 &  7.9314\\
\hline
\end{tabular}
\label{table:cases}
\end{center}
\end{table}
In total, six simulation cases are displayed in this section, see Table~\ref{table:cases}. The control outputs of Case 3 are shown in Fig. \ref{fig:60_10} as an example. Small steady state errors are accepted which do not effect the general performance of the aerial manipulator for the required manipulation tasks. The oscillation at the beginning indicates the phase when the UAM starts the physical interaction with the work surface, then it stabilizes and maintains a stable contact position and orientation while generating a consistent interaction force to the environment as required by the tasks. The positive definite gains acting on attitude control, vertical position control, and interaction force control in the proposed interaction controller are properly tuned based on study objectives.

\begin{figure}[t]
    \centering
      \includegraphics[width=\columnwidth]{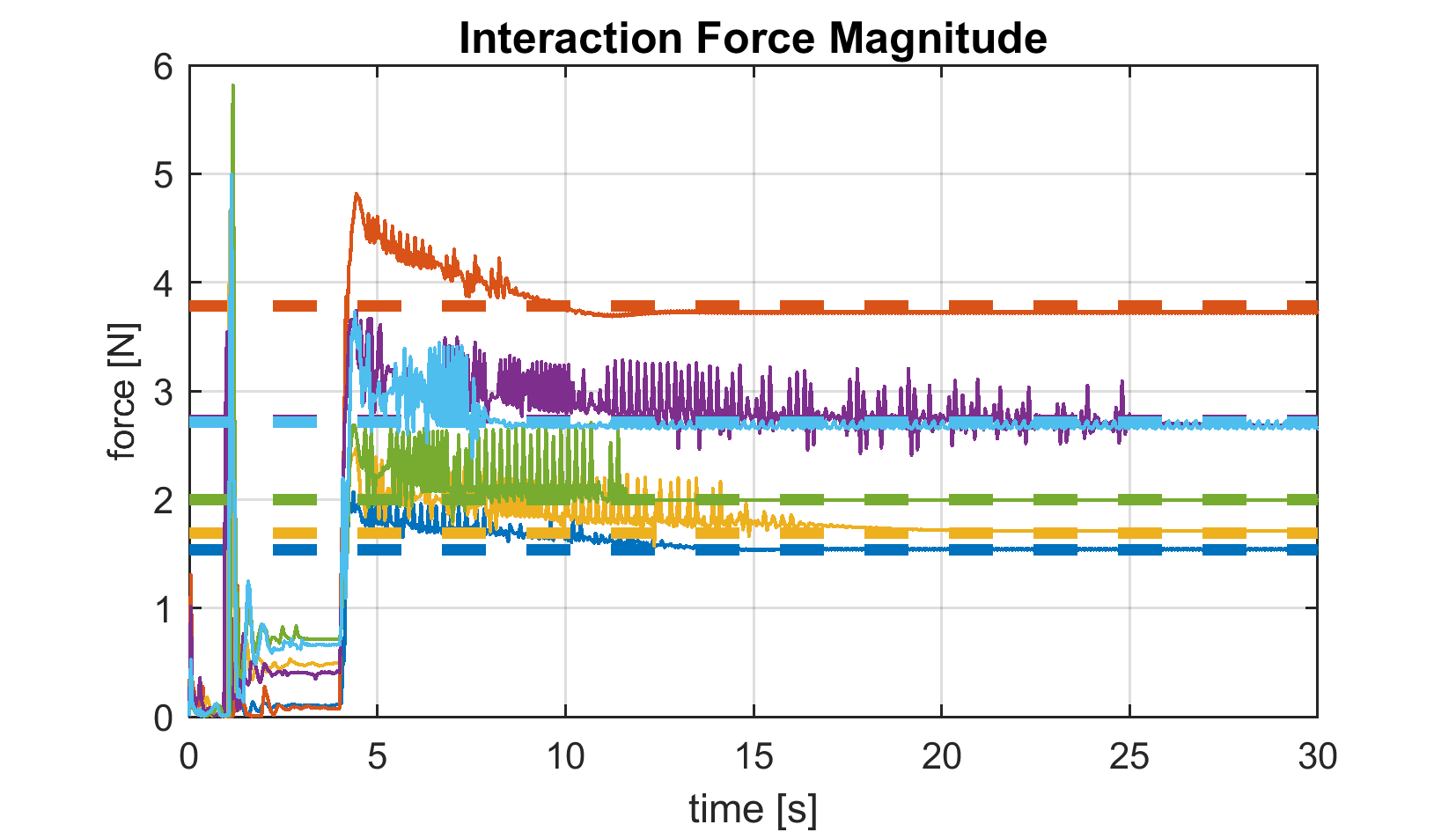}
      \caption{dashed line: desired $f_{C,Z_E}$; solid line: estimated $f_{C,Z_E}$}
      \label{fig:fe}
\end{figure}

\begin{figure}
      \includegraphics[width=\columnwidth]{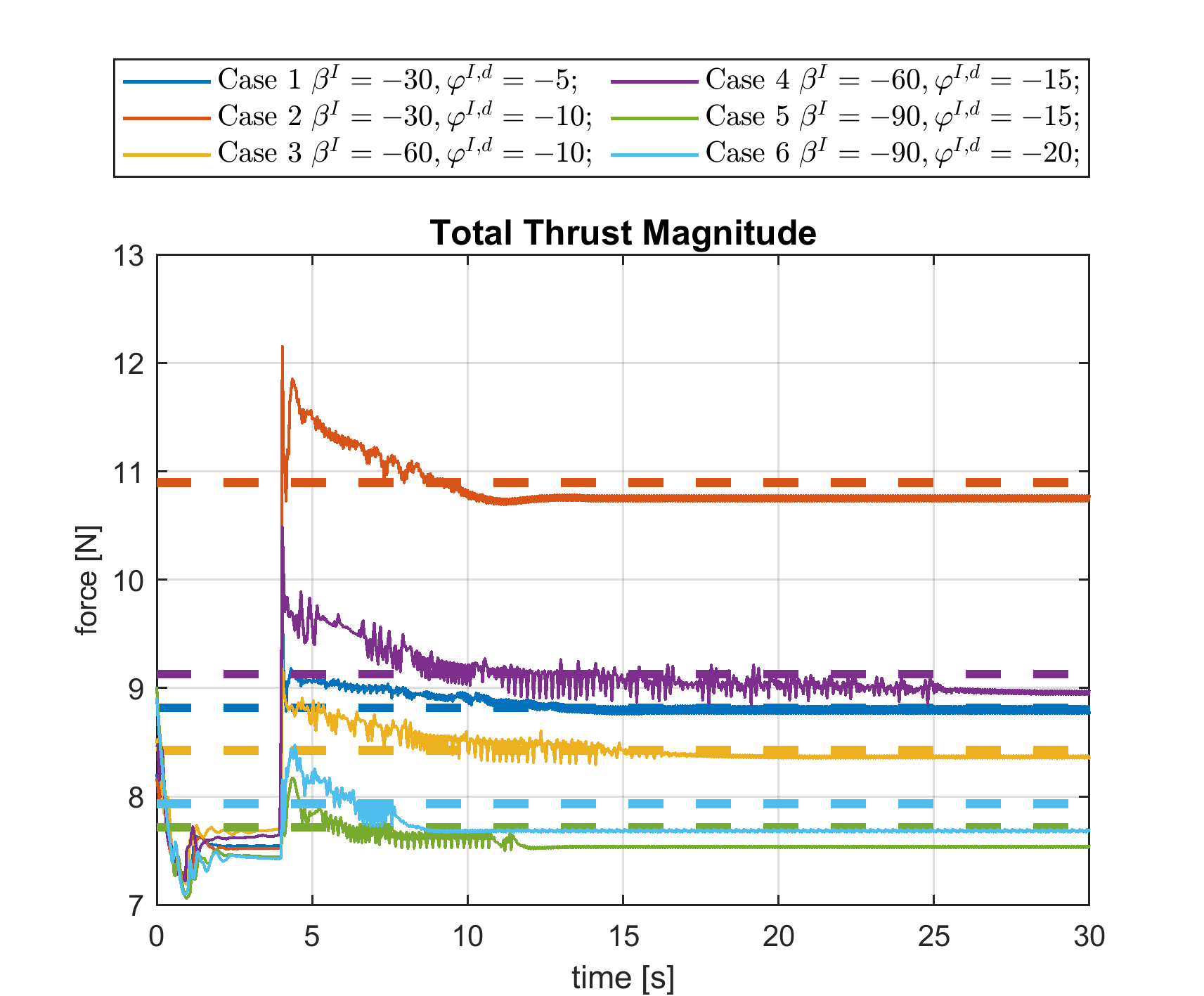}
      \caption{dashed line: desired $T_{sum}^{int}$; solid line: measured $T_{sum}^{int}$}
      \label{fig:tsum}
\end{figure}

The force profiles in Fig.~\ref{fig:plotf} are verified by both vertical and horizontal comparison among the six cases, see Table~\ref{table:cases} and Fig. \ref{fig:fe}, \ref{fig:tsum}. The vertical comparison via \textbf{Cases} 2\&3 and \textbf{Cases} 4\&5 shows that the exerted interaction force magnitude along the desired direction (as well as the demanded total thrust magnitude) is larger for a smaller $|\beta^{\mathbb{I}}|$ value defined work surface while having the same reference roll angle magnitude $|\varphi^{\mathbb{I},d}|$ of the aerial vehicle. \textbf{Cases} 1\&2, \textbf{Cases} 3\&4, and \textbf{Cases} 5\&6 instead allow a horizontal comparison in Fig. \ref{fig:plotf}, which indicates that with the same environment parameter $|\beta^{\mathbb{I}}|$, the contact force magnitude exerted along the desired direction increases along with the bigger reference roll angle magnitude $|\varphi^{\mathbb{I},d}|$ (i.e., smaller reference joint position magnitude $|\alpha^{\mathbb{I},d}|$) during the interaction. Moreover, by increasing the desired UAV roll angle magnitude $|\varphi^{\mathbb{I},d}|$ with a value of 5 degree, as shown in the horizontal comparison cases, the increment of the interaction force magnitude/total thrust magnitude is bigger for smaller $|\beta^{\mathbb{I}}|$ value.
\begin{figure}[t]
    \centering
      \includegraphics[width=\columnwidth]{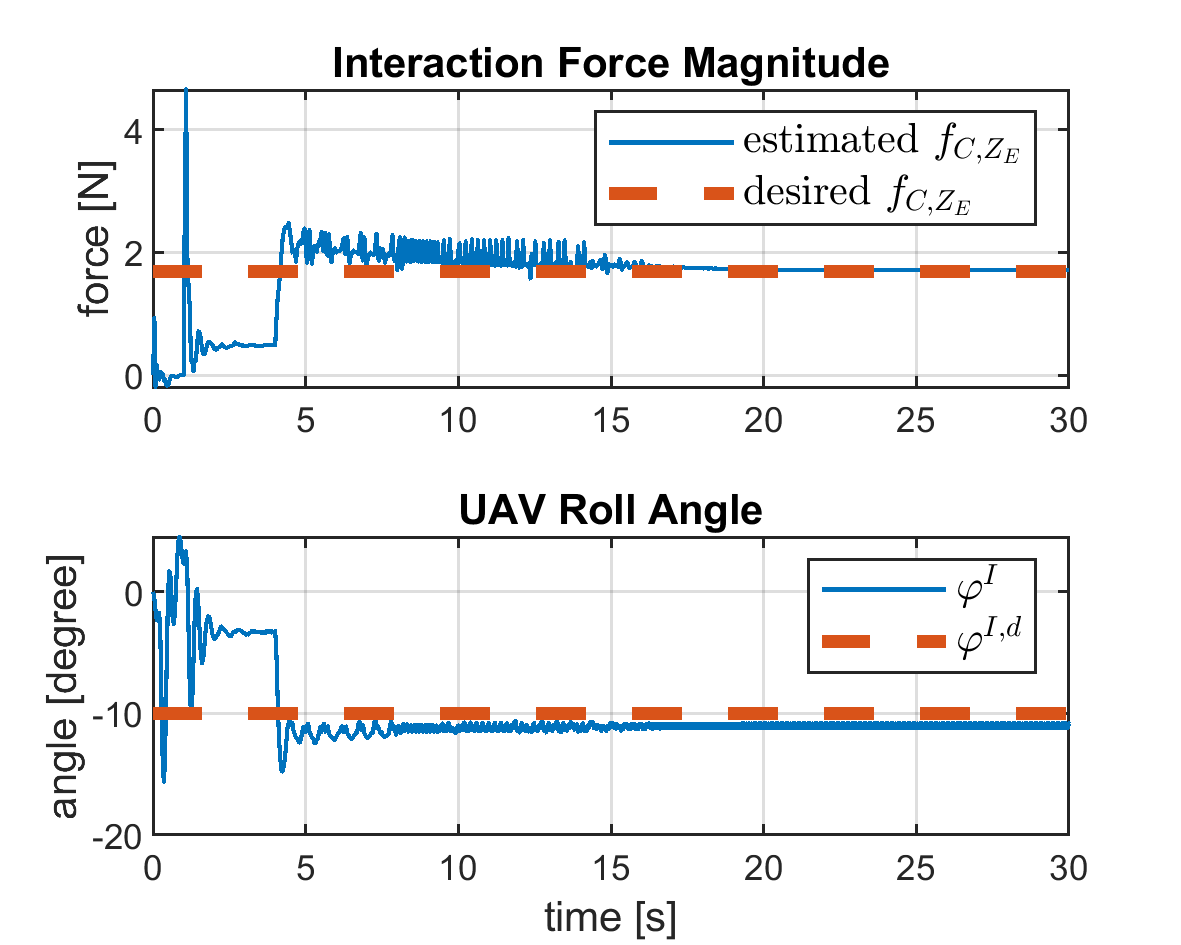}
\end{figure}

In summary, when interacting with the environment defined by a smaller $|\beta^{\mathbb{I}}|$ value (considering both negative and positive $\beta^{\mathbb{I}}$), it is more critical for the UAM to complete the tasks in Sec.~\ref{sec:ps} since it has a smaller operating range of roll angle magnitude $\in [0,|\beta^{\mathbb{I}}|)$. Especially, even a small variation of roll angle magnitude causes an evident change of the exerted interaction force and corresponded total thrust magnitude, which might lead to system saturation before the system could converge to the desired stationary interaction state. To the best of the author's knowledge, this paper firstly explores the different physical interaction conditions utilizing UDT-based aerial manipulators with variously orientated work surfaces, and introduces possible critical manipulation cases and singularity.

\section{Conclusion}
\label{sec:con}
This paper studies the feasibility of employing UDT multirotors for aerial manipulation purpose towards industrial needs. With the interest of meaningful industrial applications with UDT-based aerial manipulation and exploring the properties of such systems due to underactuation, the paper proposes a range of manipulation tasks under variant environment conditions, with which a static-equilibrium analysis based interaction force modeling approach is introduced. The paper points out the singularity during physical interaction with various work surfaces based on the explicit interaction force model in relation with the UAV attitude and the environment parameter. Moreover, a hybrid attitude/force control strategy is designed based on the modeling approach. The aforementioned interaction force modeling and control framework is then verified via simulations. This is a preliminary work. In the future work, physical experiments are planned, especially to explore the singularity or critical operating cases introduced. More complicated manipulation tasks which require interaction rotational torques and a more general force modeling approach in three dimensional space will be studied. Investigation in trajectory and motion planning for the path of approaching work surface starting from a free flight point is needed to have more accurate contact point position. Moreover, manipulator dynamics and actuation will be included in the system model and control design.

\addtolength{\textheight}{-12cm}   






% ...

\end{document}